\pdfoutput=1

\documentclass[11pt]{article}
\usepackage[table]{xcolor}
\usepackage{acl}
\usepackage{adjustbox}
\usepackage{tikz}
\usepackage{times}
\usepackage{graphicx}
\usepackage{booktabs}
\usepackage{latexsym}
\usepackage{multirow}
\usepackage{amssymb}
\usepackage{amsmath}
\usepackage{pifont}
\usepackage{tabularx}
\usepackage{marginnote}
\usepackage{cleveref}
\usepackage[T1]{fontenc}

\usepackage[utf8]{inputenc}

\usepackage{microtype}

\crefformat{section}{\S#2#1#3}
\crefformat{subsection}{\S#2#1#3}
\crefformat{subsubsection}{\S#2#1#3}

\title{Low-Resource Dense Retrieval for Open-Domain Question Answering: \\A Comprehensive Survey}

\author{Xiaoyu Shen$^{\dagger{1}}$, Svitlana Vakulenko$^{1}$, Marco del Tredici$^{1}$, Gianni Barlacchi$^{1}$\\ \textbf{Bill Byrne}$^{1,2}$  \textbf{and Adrià de Gispert}$^{1}$ \\
$^1$Amazon Alexa AI\\
$^2$Univeristy of Cambridge \\
$\dagger$\tt{gyouu@amazon.com}}

\begin{document}
\maketitle
\begin{abstract}
Dense retrieval (DR) approaches based on powerful pre-trained language models (PLMs) achieved significant advances and have become a key component for modern open-domain question-answering systems. However, they require large amounts of manual annotations to perform competitively, which is infeasible to scale. To address this, a growing body of research works have recently focused on improving DR performance under low-resource scenarios. These works differ in what resources they require for training and employ a diverse set of 
techniques. Understanding such differences is crucial for choosing the right technique under a specific low-resource scenario. To facilitate this understanding, we provide a thorough structured overview of mainstream techniques for low-resource DR. Based on their required resources, we divide the techniques into three main categories: (1) only documents are needed; (2) documents and questions are needed; and (3) documents and question-answer pairs are needed. For every technique, we introduce its general-form algorithm, highlight the open issues and pros and cons. Promising directions are outlined for future research.
\end{abstract}

\section{Introduction}
Open-Domain Question Answering (ODQA) aims to provide precise
answers in response to the user’s questions by drawing on a large collection of documents~\cite{voorhees1999trec}.
The majority of the modern ODQA systems follow the retrieve-and-read architecture: 1) given a question, a set of relevant documents are retrieved from a large document collection, and 2) the reader-model produces an answer given this set~\cite{chen2017reading,verga2021adaptable,lee2021you,RaposoRMC22}.

Conventional methods use sparse retrievers (SRs) such as TF-IDF and BM-25 in the retrieving stage to match questions and documents via lexical overlap. These can be considered as representing the question and document as sparse bag-of-word vectors~\cite{robertson1994some,robertson2009probabilistic}. SRs have an important limitation: they may overlook semantically relevant documents that have low lexical overlap with the question~\cite{chowdhury2010introduction}.
Dense-retrievers (DRs) resolve this issue by encoding the questions and documents into dense vectors so that synonyms and paraphrases can be mapped to similar vectors through task-specific fine-tuning~\cite{das2019multi,karpukhin2020dense}.
However, DRs require large amounts of annotated text to perform competitively and have also been found to generalize poorly across domains~\cite{thakur2021beir,ren2022thorough}. In practice, collecting annotations for question-document relevance is very time-consuming. For a given question, an extensive annotation effort may be required to find the relevant documents. Repeating this annotation process for every new 
language and domain is not feasible.

\begin{table*}
\centering
\small
\begin{tabularx}{\textwidth}{l|l|l|c}
\toprule
\textbf{Resource} & \textbf{Technique} & \textbf{Strategy} & \textbf{OOD Required}\\
\hline
\multirow{3}{*}{\shortstack[l]{Documents (\cref{sec:documents})}} & Denoising Auto Encoding (\cref{sec:dae}) & Learn better in-domain representations & No\\
& \cellcolor{gray!25}Self Contrastive Learning (\cref{sec:scl}) & \cellcolor{gray!25}Build pseudo question-document pairs& \cellcolor{gray!25}No \\
& \cellcolor{gray!25}Question Generation (\cref{sec:qg}) & \cellcolor{gray!25}Build pseudo question-document pairs& \cellcolor{gray!25}Optional\\
\hline
\multirow{2}{*}{\shortstack[l]{ Documents\\+Questions (\cref{sec:questions})}} &  \cellcolor{gray!25}Distant Supervision (\cref{sec:ds}) & \cellcolor{gray!25}Build pseudo question-document pairs & \cellcolor{gray!25}Optional\\
&  Domain-Invariant Learning (\cref{sec:dat}) & Learn domain-invariant representations & Yes\\
\hline
\multirow{3}{*}{\shortstack[l]{Documents\\+QA Pairs (\cref{sec:qa_pairs})}} &  \cellcolor{gray!25}Direct fine-tuning (\cref{sec:df}) & \cellcolor{gray!25}Build pseudo question-document pairs & \cellcolor{gray!25}No\\
&  \cellcolor{gray!25}Answer-Document Mapping (\cref{sec:adm}) & \cellcolor{gray!25}Build pseudo question-document pairs & \cellcolor{gray!25}No\\
&  Latent-Variable Model (\cref{sec:lvm}) & Jointly train the retriever and reader & No\\
\hline
\multirow{3}{*}{\shortstack[l]{Others (\cref{sec:others})}} &  Integration with Sparse Retrieval & Combine advantages of DR and SR & Optional\\
&  Multi-Task Learning & Improve generalization from multiple tasks & Optional\\
&  Parameter Efficient Learning & fine-tune fewer parameters to reduce overfit & Optional\\
\hline
\toprule
\end{tabularx}
\caption{\label{tab:techniques}
\small Overview of different techniques grouped by the in-domain resources they need. ``\textbf{OOD required}'' indicates if the corresponding technique further requires out-of-domain annotations. \colorbox{gray!25}{Techniques with gray background} adopt the same idea: build pseudo question-document pairs and train the model on the pseudo pairs.
}
\end{table*}

Techniques have recently been proposed to improving DR training in low-resource scenarios, i.e., when \emph{only a few or no (zero-shot) question-document annotations are available for the target domain}~\cite{reddy2021towards,wang2022gpl}.
These techniques differ in what resources they require for training and employ a diverse set of algorithms. Due to the large differences inherent in different domains, most techniques require some \emph{in-domain} resource for adequate performance.
We begin by providing an overview of common in-domain resources, beyond question-document annotations, that can be used for training DRs:
\begin{enumerate}
    \item \textbf{Documents}: a document corpora is a bare minimum for building the retriever, which can be leveraged for improving DRs.
    \item \textbf{Questions}: a set of sample questions may be available but without ground-truth document or answer annotations. 
    \item \textbf{QA Pairs}: question-answer pairs can exist in some domains with a large number of already answered questions, e.g., in customer service, technical support or web forums. 
\end{enumerate}

For each of the resources, we review applicable techniques and explain how they can improve the retrieving performance with limited annotated data. Some techniques have additional requirements such as manually defined heuristics or out-of-domain (OOD) annotations, which we will also discuss in the corresponding sections. An overview of techniques can be seen in Table~\ref{tab:techniques}. Structuring techniques by the resources they require allows us to organise this survey as a practical guide for choosing the best techniques given the available resources. 

In the following sections we provide an overview of related work (\cref{sec:related_work}) and lay out the background (\cref{sec:background}) necessary for a more in-depth understanding of the techniques we discuss in~\cref{sec:documents,sec:questions,sec:qa_pairs,sec:others}.
In conclusion, we highlight open issues and promising directions for future work (\cref{sec:conclusions}).

\section{Related Work}
\label{sec:related_work}
This survey contributes to the growing body of work describing the state-of-the-art approaches to QA~\cite{zeng2020survey,zhu2021retrieving,roy2021question,rogers2021qa,pandya2021question}.
Unlike previous surveys that describe general neural information retrieval (IR) approaches~\cite{mitra2018introduction,guo2020deep,lin2021pretrained,guo2022semantic}, we focus specifically on the low-resource scenarios, which makes our contribution unique in this respect.
The closest to our work is the BEIR benchmark for zero-shot cross-domain evaluation of IR models~\cite{thakur2021beir}
and its multiple related studies~\cite{mokrii2021systematic,reddy2021towards,wang2022gpl,ren2022thorough}.
These studies test specific algorithms but do not provide a holistic overview of how they are related.
For further details on each of the techniques described here, readers can refer to relevant in-depth reviews, e.g., for question generation~\cite{zhang2021review}, knowledge distillation~\cite{gou2021knowledge} and sentence representation learning~\cite{li2022brief}.

\begin{figure*} 
\begin{center} 
\includegraphics[width=0.9\textwidth]{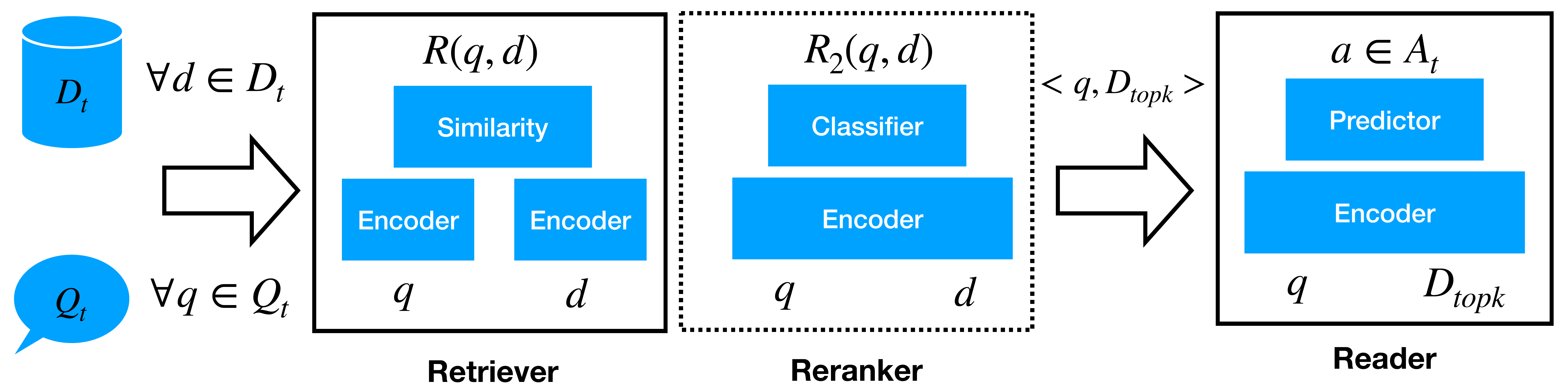}
\caption{\small A typical architecture of an open-domain QA system. It consists of an initial retriever and an optional reranker to produce top-$k$ relevant documents $D_{topk}$ given a question $q$, and a reader to predict the final answer based on $q$ and $D_{topk}$. The retriever usually adopts the bi-encoder architecture to encode questions and documents separately, while the reranker can afford a cross-encoder architecture to allow full question-document interactions.} 
\label{fig:arch} 
\end{center} 
\end{figure*}

\section{Background}
\label{sec:background}
We begin with a brief introduction to dense retrieval for ODQA and address the definition of ``domain'' as used in this survey. We finally elaborate on the low-resource scenario that we will be focusing on.
\paragraph{Dense Retrieval for ODQA}
Let $Q_t, D_t$ and $A_t$ denote the question, document~\footnote{We use the term ``document'' for the retrieval unit. Depending on the granularity in the task, it can refer to a document, passage, sentence, table, etc~\cite{hofstatter2020fine}.} and answer set from the target domain $t$. Given a question $q \in Q_t$, for each $d \in D_t$, the retriever of the ODQA system assigns a relevance score $R(q,d)$ to them. For computational efficiency, normally a single-representation-based bi-encoder architecture is applied. In a bi-encoder architecture, the question and document are encoded independently into two vectors $E1(q)$ and $E2(d)$~\cite{bromley1993signature}, where $E1$ and $E2$ are the question and document encoders. $E1$ may share the same parameters with $E2$ and their dot-product or cosine similarity is used as the relevance score~\cite{reimers2019sentence,xiong2020approximate}. Apart from the bi-encoder architecture, we can also use multiple representation retrieval such as Colbert~\cite{khattab2020colbert}. Multiple representation retrieval learns relevance scores based on interactions multiple representations from questions and documents. This can improve the model capacity but reduce the inference efficiency. For every question $q$, the top-$k$ documents with the highest relevance scores will be returned~\footnote{Due to the computational complexity, normally nearest neighbor search tools like FAISS~\cite{johnson2019billion} are used to obtain the top-$k$ documents with sublinear complexity.}. The training objective for the retriever can be formalized as:
\begin{align}
\begin{split}
      \max_{R}\mathbb{E}_{q,d^{+}, d_{1\sim n}^{-}\in Q_t\times  D_t}\mathcal{O}(R, q, d^{+}, d_n^{-})
\end{split}
\label{eq:dr_objective}
\end{align}
where $Q_t \times D_t$ indicates the full set of question-answer pairs, $d^+$ is a positive (relevant) document for $q$, $d_{1\sim n}^{-}$ is the sampled $n$ negative (irrelevant) documents and $\mathcal{O}$ is the objective function. A common choice for $\mathcal{O}$ is the contrastive objective:
\begin{align}
\begin{split}
    \mathcal{O}=&\log\frac{e^{R(q,d^{+})}}{e^{R(q,d^{+})}+\sum_{j=1}^{n}e^{R(q,d_j^{-})}}
\end{split}
\end{align}
After retrieval, a reranker can optionally be applied to rerank the retrieved results. As it is only applied to the top retrieved documents, we can afford to use a more powerful architecture. A typical approach is to concatenate the question and document and then use a cross-encoder classifier to get a score $R_2(q,d)$ for every question-document pair found in the initial retrieval. The cross-encoder architecture allows full interaction between the questions and documents so that the relevance scores are better estimated. Finally, a reader will estimate the score $G(a|q,D_{topk})$ to generate the final answer $a \in A_t$ conditioned on both the question and the top-$k$ retrieved documents $D_{topk} \subset D_t$, as ranked by the relevance scores produced by the retriever or reranker. Figure~\ref{fig:arch} illustrates this process.
\paragraph{Definition of Domain} Typically in NLP, ``domain'' refers to a coherent type of corpus, i.e., predetermined by a given dataset~\cite{plank2011domain}. The type may
relate to various dimensions of latent factors like topic (e.g., sports, finance, biology), genre (e.g., social media, newswire, scientific), style (e.g., formal/informal, male/female) and language (e.g., English, Spanish, Chinese)~\cite{ramponi2020neural}. Strictly speaking, no two corpora can be exactly the same in all their factors. Change in language use and emergence of topics over time will also yield new domains. Therefore, enumerating over all possible domains is impossible and systems are best developed with a particular task in mind. We use the term ``in-domain'' to refer to data that comes from the same type of corpus as our task, whereas ``out-of-domain'' (OOD) refers to data that comes from another type of corpus that can differ from our target task in one or several factors (e.g., different topics, genres, styles or languages). We take a broader definition by considering language as one of the factors that defines a domain since every domain could have its domain-specific language~\cite{hedderich2021survey}. Although techniques summarized in this survey can in principle be applied to 
arbitrary language, we do not cover two specific scenarios: (1) \emph{the questions and documents are in different languages}, and (2) \emph{the OOD annotated data is in a different language than the target domain}. In these scenarios, cross-lingual transfer techniques are needed, which is out of the scope of this survey. Interested readers can refer to \citet{pikuliak2021cross,litschko2022cross} for more information on this topic.

\begin{figure*} 
\begin{center} 
\includegraphics[width=\textwidth]{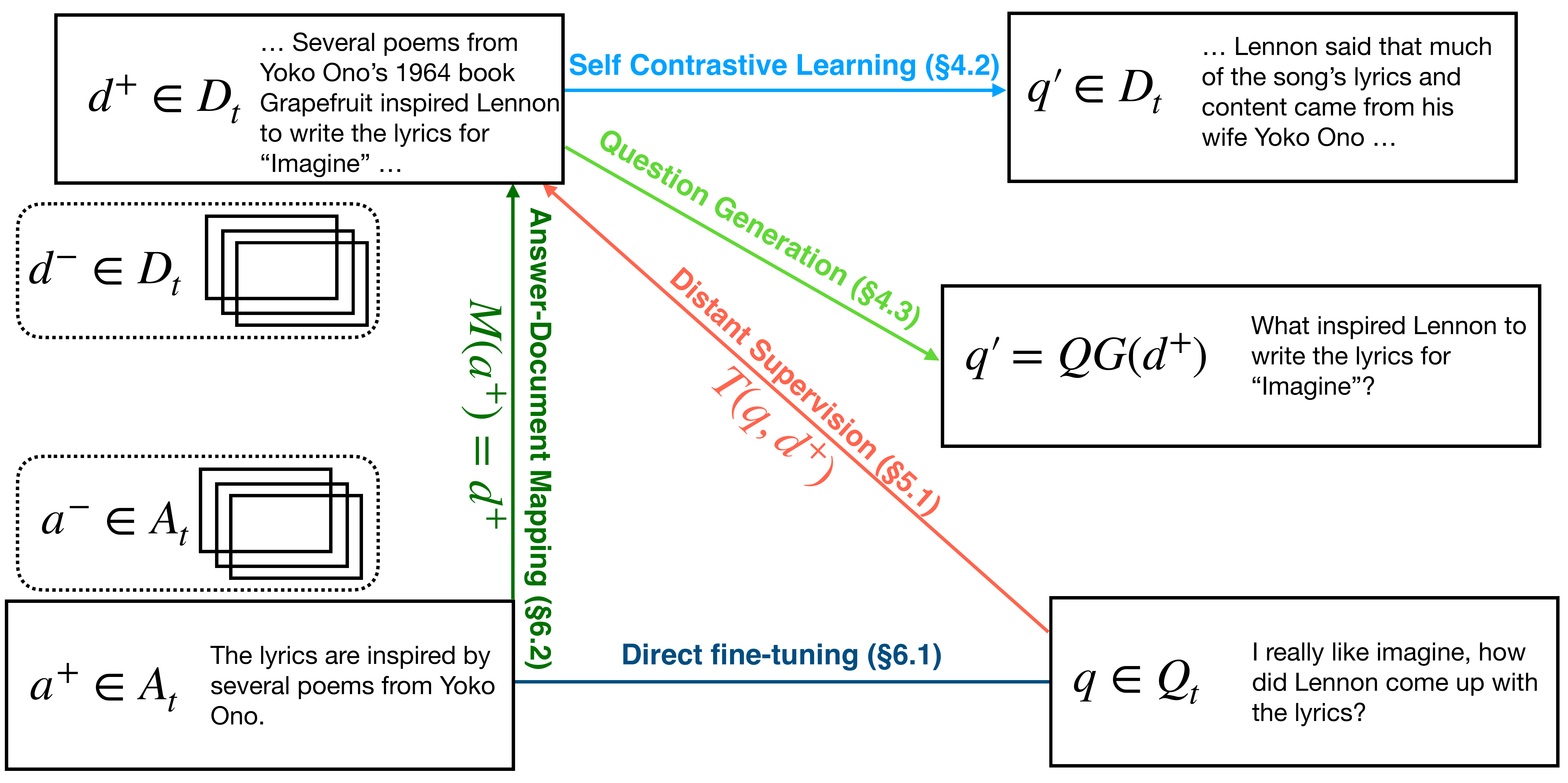}
\caption{\small Five different techniques of building pseudo question-document pairs. Each line refers to one technique and boxes indicate the required resources. Given a document $d^+\in D_t$, self-contrastive learning finds a pseudo question $q^\prime\in D_t$ while question generation generates a new question $q^\prime=QG(d^+)$. Distant supervision applies a teacher model $T$ to link each unlabeled question $q\in Q_t$ to existing documents via $T(q,d^+)$. Direct fine-tuning treats the answer $a^+$ as the positive document, while answer-document mapping maps the answer $a^+$ into an existing document $M(a^+)=d^+$.} 
\label{fig:pseudo_qa} 
\end{center} 
\end{figure*}

\paragraph{Low-resource Scenario}
In the standard supervised setting we need annotations for $(q_t,d_t) \rightarrow \{+,-\}$ to train the retriever with Eq~\ref{eq:dr_objective}. Obtaining such annotations requires tremendous human labor and is expensive to scale to multiple domains~\cite{ram2021learning}. We focus on the low-resource scenario where either zero or only a few annotations are available in the target domain. To offset the lack of annotated data, we need to make full use of whatever in-domain resources are available to train the retriever. We group techniques by 3 common in-domain resources that we can leverage: (1) \emph{Documents}: document collections $D_t$; (2) \emph{Documents + Questions}: document collection $D_t$ and question set $Q_t$; (3) \emph{Documents + QA Pairs}, document collection $D_t$ and QA pairs $(Q_t, A_t)$. An overview of the techniques can be seen in Table~\ref{tab:techniques}~\footnote{Even though we focus on training the dense retrieval in this survey, most techniques are architecture-agnostic and can be straightforwardly applied to training the reranker as well.}. Apart from these in-domain resources, we can also leverage OOD annotations $(Q_o,D_o) \rightarrow \{+,-\}$ where $(Q_o,D_o)$ are question-document pairs from a different domain. We note in the corresponding sections if OOD annotations are needed. Most techniques follow the same idea: build pseudo question-document pairs then train the model on them. Figure~\ref{fig:pseudo_qa} illustrates various strategies to build such pseudo pairs.
In the next section, we will present the three in-domain resources and discuss the techniques for which they are applicable.

\section{Leveraging Documents}
\label{sec:documents}
This section discusses techniques that exploit in-domain documents to train the DR. This makes the minimum assumption about resource availability since having in-domain documents is a prerequisite for building a retriever. 

There are three popular techniques of leveraging documents: (1) denoising auto encoding, (2) self contrastive learning, and (3) question generation.

\subsection{Denoising Auto Encoding}
\label{sec:dae}
 Denoising auto encoding (DAE)~\cite{vincent2010stacked} is a common approach for representation learning. The model contains an encoder $E$ and decoder $F$. $E$ takes as input a corrupted sentence $\tilde{s}$ from $D_t$, and $F$ learns to reconstruct the original sentence $s$. The training objective is:
\begin{align}
    \min_{E,F}\mathbb{E}_{s\in D_t} \mathcal{L}(F(E(\tilde{s})),s)
\end{align}
where $\mathcal{L}$ is a loss function that encourages a faithful reconstruction; typically the cross entropy loss is used. The corrupted sentence $\tilde{s}$ can be obtained by adding noise such as word masking, deletion and shuffling, to $s$. After training, the encoder $E$ should learn rich in-domain semantic representations and map similar sentences to similar vector representations~\cite{gururangan2020don,karouzos2021udalm}. We can then use the learnt encoder $E$ to initialize the retriever. There are two common forms of auto encoding depending on the choice of the decoder $F$: Enc-Only and Enc-Dec.

\paragraph{Enc-Only} The Enc-only form does not have a separate decoder but reuses the encoder for reconstruction. A notable example is masked language modelling (MLM) as is commonly used for pretrained language models (PLMs)~\cite{kenton2019bert,liu2019roberta}. The original form of MLM predicts each output token (or differentiate it from other tokens~\cite{clark2020electra}) independently based on the encoded representation at each masked position. \citet{gao2021condenser} found this token-level representation is not suitable for bi-encoders, which aggregate text information into the dense representation. They propose to condition the MLM prediction explicitly on dense representations and demonstrate improvement over conventional MLM models. The Enc-only architecture is simple and allows full parameter sharing between encoding and reconstruction, but is less flexible, in that other forms of decoder architectures, such as autoregressive, cannot be used~\cite{wang2019bert}. Enc-only architectures also requires a pre-determined number of decoded tokens. Special tricks such as padding are needed if corrupted sentences have different number of tokens than the original~\cite{malmi2020unsupervised}.

\paragraph{Enc-Dec} The Enc-Dec architecture applies a separate decoder for reconstruction. It is often preferred over the Enc-only architecture since this split allows choosing other forms of decoders for specific use cases~\cite{tay2022unifying}. The most popular choice are autoregressive decoders where each output token is predicted conditioned on previous tokens, an approach that achieves state-of-the-art performance in generative tasks~\cite{lewis2020bart,raffel2020exploring,zhang2020pegasus}. For retrieval tasks, as we only care about the encoder, a weaker decoder is often used to encourage richer information to be included in the encoded representations. The decoder can be made weaker by various ways such as limiting the size of context~\cite{yang2017improved}, tied parameters and constrained attention~\cite{wang2021tsdae}, reducing the decoder size~\cite{lu2021less}, or polluting tokens on the decoder side~\cite{liu2022retromae}.

\paragraph{Discussion} DAE-based methods are easy to implement and train efficiently. However, the auto-encoding objective is very different from the contrastive objective for training the DR. Since no further regularization is added to the encoder, the learnt representation space can often collapse so that representations cannot be easily distinguished~\cite{chen2021exploring,yan2021consert}. One possible solution is to specify prior distributions to regularize the representation space as with variational autoencoders~\cite{kingma2014auto,li2020optimus}, but this requires implement tricks and careful tuning to avoid the ``posterior collapse''~\cite{alemi2018fixing}. Another drawback is that it cannot be readily applied if the question and document use two different encoders, since DAE only produces a single encoder. In practice, DAE is often used to initialize the encoder parameters, but performance is poor without further fine-tuning on labeled data~\cite{dai2019deeper,wang2022gpl}.
\begin{table}
\centering
\small
\begin{tabularx}{0.46\textwidth}{c|X}
\toprule
\textbf{Method} & \textbf{Pseudo Question} \\
\hline
\multirow{2}{*}{\shortstack[c]{Perturbation\\based}} & \multirow{2}{*}{\shortstack[c]{$d$ with added perturbation}}\\
&\\
\hline
\multirow{2}{*}{\shortstack[c]{Summary\\based}} & \multirow{2}{*}{\shortstack[c]{(Pseudo) summary of $d$}}\\
&\\
\hline
\multirow{2}{*}{\shortstack[c]{Proximity\\based}} & \multirow{2}{*}{\shortstack[c]{Nearby text of $d$}}\\
&\\
\hline
\multirow{2}{*}{\shortstack[c]{Cooccurrence\\based}} & \multirow{2}{*}{\shortstack[c]{Text sharing cooccurred spans with $d$}}\\
&\\
\hline
\multirow{2}{*}{\shortstack[c]{Hyperlink\\based}} & \multirow{2}{*}{\shortstack[c]{Text with a hyperlink to/from $d$}}\\
&\\
\toprule
\end{tabularx}
\caption{\label{tab:sct}
\small Given a text $d$ as the pseudo document, different ways of constructing pseudo questions for it in \textbf{self contrastive learning}. These pseudo questions are paired with $d $ to form positive training instances. Details are in \cref{sec:scl}.
}
\end{table}

\subsection{Self Contrastive Learning}
\label{sec:scl}
 \emph{Self Contrastive Learning} (SCT) applies different heuristics to construct pseudo question-document pairs $(q^\prime,d^{\prime+/-})$ from $D_t$, then train the model to maximize the scores of positive pairs and minimize the scores of negative ones. By this means, the training objective can be consistent with the loss used in retrieving tasks. The training objective of SCT is:
\begin{align}
\begin{split}
     \max_{R}\mathbb{E}_{q^\prime,d^{\prime+}, d_{1\sim n}^{\prime-}\in D_t}\mathcal{O}(R, q^\prime,d^{\prime+},d_{1\sim n}^{\prime-})
\end{split}
\end{align}
where $\mathcal{O}$ is the standard ranking objective as in Eq~\ref{eq:dr_objective}.
Since negative pairs can be easily constructed by random sampling, the main difficulty in SCT is to design good heuristics for constructing positive pseudo pairs $(q^\prime,d^{\prime+})$.
There are 5 popular ways of constructing such positive pairs: perturbation-based, summary-based, proximity-based, cooccurence-based and hyperlink-based. An overview is in Table~\ref{tab:sct}.

\paragraph{Perturbation-based} Perturbation-based methods add perturbations to some text, then treat the perturbed text and the original text as a positive pair. The intuition is that \emph{a text, after applying some perturbation on it, should still be relevant with the original text}. Typical choices of perturbations include word deletion, substitution and permutation~\cite{zhu2021contrastive,meng2021coco}, adding drop out to representation layers~\cite{gao2021simcse}, or passing sentences through different language models~\cite{carlsson2021semantic}, among other. 

\paragraph{Summary-based}  Summary-based methods extract a summary from the document as the pseudo question based on the intuition that \emph{questions should contain representative information about the central topic of the document}. The summary can be the document title~\cite{macavaney2017approach,macavaney2019content,mass2020ad}, a random sentence from the first section of the document~\cite{chang2020pre}, randomly sampled ngrams~\cite{gysel2018neural} or a set of keywords generated from a document language model~\cite{ma2021prop}.

\paragraph{Proximity-based}  Proximity-based methods utilize the position information in the document to obtain positive pairs based on the intuition that \emph{nearby text should be more relevant to each other}. The most famous one is the inverse-cloze task~\cite{lee2019latent}, where a sentence from a passage is treated as the question and the original passage, after removing the sentence, is treated as a positive document. They can be combined with typical noise injection methods like adding drop-out masks~\cite{xu2022laprador}, random word chopping or deletion~\cite{izacard2021towards} to further improve the model robustness. Other methods include using spans from the same document~\cite{gao2022unsupervised,ma2022pre}, sentences from the same paragraph, paragraphs from the same documents as positive samples~\cite{di2022pre}, etc.

\paragraph{Cooccurrence-based}  Cooccurrence-based methods construct positive samples based on the intuition that \emph{sentences containing cooccurred spans are more likely to be relevant}~\cite{ram-etal-2021-shot}. For example, \citet{glass2020span} constructs a pseudo question with a sentence from the corpus. A term from it is treated as the answer and replaced with a special token. Passages retrieved with BM25 which also contains the answer term are treated as pseudo positive documents. \citet{ram2021learning} treated a span and its surrounding context as the pseudo question and use another passage that contains the same span as a positive document. 

\paragraph{Hyperlink-based}  Hyperlink-based methods leverage hyperlink information based on the intuition that \emph{hyperlinked texts are more likely to be relevant}. For example, \citet{zhang2020selective,ma2021pre} revisited the classic IR intuition that anchor-document relations approximate question-document relevance and build pseudo positive pairs based on these relations. \citet{chang2020pre} takes a sentence from the first section of a page $p$ as a pseudo question because it is often the description or summary of the topic. A passage from another page containing hyperlinks to $p$ is treated as a positive document. \citet{yue2022c} replaced an entity word with a question phrase like ``what/when'' to form a pseudo question. A passage from its hyperlinked document that contains the same entity word is treated as a positive sample. \citet{zhou2022hyperlink} builds positive samples with two typologies: ``dual-link'' where two passages have hyperlinks pointed to each other, and ``co-mention'' where two passages both have a hyperlink to the same third-party document.

\paragraph{Discussion}
SCT usually have much better zeroshot performance than DAE since the encoders are explicitly trained with discriminative objectives. However, previous research has shown that the benefits of SCT are wiped out after being fine-tuned on labeled data, even if the data is out of domain~\cite{liu2022retromae}. It is unclear to which extent SCT is helpful when additional labeled data is available. The training is also less memory-efficient and requires large numbers of negative samples for every batch~\cite{gao2022unsupervised}. Furthermore, it is non-trivial to decide which heuristics to apply when we face a new domain. One possible solution is to automatically select good pseudo pairs with reinforcement learning (RL) when minimal target-domain annotations are available~\cite{zhang2020selective}, but it would further slow down and complicate the training process.
In practice hyperlink-based approaches usually perform the best as they have additional reference information to leverage, which makes them most similar to the actual relevance annotations. However, hyperlink information is not available in most domains and thereby limits its use cases~\cite{sun2021few}.

\subsection{Question Generation}
\label{sec:qg}
 DAE and SCT rely solely on sentences already present in $D_t$. \emph{Question generation} (QG) constructs pseudo training pairs by generating new questions \emph{not found} in $D_t$. This approach employs a question generator $QG$ and, very often, a filter $Fil$. $QG$ can generate questions conditioned on the document while $Fil$ is used to filter poorly generated questions. The training objective is:
\begin{align*}
\begin{split}
     \max_{R}\mathbb{E}_{q=QG(d^{+}) \& Fil(q,d^{+})=0}
     \mathcal{O}(R,q,d^{+},d_{1\sim n}^{-})
\end{split}
\end{align*}
where the expectation is with respect to documents $d^+$ and $d_{1\sim n}^{-}$ drawn from $D_t$, the $q$ are generated from $d^+$, $Fil(q,d^{+})=0$ requires that these questions not be discarded by $Fil$, and $\mathcal{O}$ is the standard ranking objective. There are various ways of designing the question generator and filter. We will cover the popular choices in the following section. An overview can be seen in Table~\ref{tab:qg}.

\begin{table}
\centering
\begin{adjustbox}{max width=0.47\textwidth}
\small
\begin{tabular}{l|l}
\toprule
\multirow{4}{*}{Inputs} &  Document \\
& Document + Answer\\
& Document + Answer + Question Type\\
& Document + Answer + Question Type + Clue\\
\midrule
\multirow{3}{*}{\shortstack[l]{Question\\Generator}}&Rule-based$\dagger$\\
& Prompt-based$\dagger$\\
& Supervised model$\ddagger$\\
\midrule
\multirow{8}{*}{Filter} &  LM Score \\
& Round-trip Consistency\\
& Probability from pretrained QA\\
& Influence Function\\
& Ensemble Consistency\\
& Entailment Score\\
& Learning to Reweight$^\diamond$\\
& Target-domain Value Estimation$^\diamond$\\
\toprule
\end{tabular}
\end{adjustbox}
\caption{\label{tab:qg}
\small Different choices for question generation. $\dagger$ means unsupervised model. $\ddagger$ means supervised model (out-of-domain annotations can be used to train it). $^\diamond$ means minimal target-domain annotations are needed.
}
\end{table}

\paragraph{Choices of Input}
A variety of information can be provided as input for the QG model. The most straightforward approach is answer-agnostic which provides only the document~\cite{du2017identifying,kumar2019cross}. In this way, the model can choose to attend to different spans of the document as potential answers and so generate different, corresponding questions. A more common method is answer-aware. An answer span is first extracted from a document, then the QG generates a question based on both the document and answer~\cite{alberti2019synthetic,shakeri2020end}. Finer-grained information can also be provided such as the question type (``what/how/...'')~\cite{cao2021controllable,gao2022makes} as well as additional clues (such as document context to disambiguate the question)~\cite{liu2020asking}. Adding more information reduces the entropy of the question and makes it easier for the model to learn, but also increases the possibility of error propagation~\cite{zhang2019addressing}. In practice, well-defined filters should be applied to remove low-quality questions.

\paragraph{Choices of Question Generator} There are three popular choices for the question generator. (1) Rule-based methods~\cite{pandey2013automatic,rakangor2015literature} rely on handcrafted templates and features. These are time-consuming to design, domain-specific, and can only cover certain forms of questions. (2) Prompt-based methods relying on PLMs. A document can be presented to a PLM, with an appended prompt such as ``Please write a question based on this passage'' so that the PLM can continue the generation to produce a question~\cite{bonifacio2022inpars,sachan2022improving,dai2022dialog}. (3) Supervised generators that are fine-tuned on question-document pairs. When in-domain annotations are not enough, we can leverage out-of-domain (OOD) annotations, if any, for QG training. The first two methods require no training data, but their quality is often inadequate. In practice, we should only consider them when there is a complete lack of high-quality supervised data for fine-tuning the QG such as when we target a domain with question shapes that are very different from any available domain.

\paragraph{Choices of Filter}
Filtering is a crucial part of QG since a significant portion of generated questions could be of low quality and would provide misleading signals when used to train the retriever~\cite{alberti2019synthetic}. A typical choice is filtering based on round-trip consistency~\cite{alberti2019synthetic,dong2019unified}, where a pretrained QA system is applied to produce an answer based on the generated question. A question is kept only when the produced answer is consistent with the answer from which the question is generated. We can also relax this strict consistency requirement and manually adjust an acceptance threshold based on the probability from the pretrained QA system~\cite{zhang2019addressing,lewis2021paq}, LM score from the generator itself~\cite{shakeri2020end,liang2020embedding}, or an entailment score from a model trained on question-context-answer pairs~\cite{liu2020asking}. Influence functions~\cite{cook1982residuals} can be used to estimate the effect on the validation loss of
including a synthetic example~\cite{yang2020generative}, but this does not achieve satisfying performances on QA tasks~\cite{bartolo2021improving}. \citet{bartolo2021improving} propose filtering questions based on ensemble consistency, where an ensemble of QA models are trained with different random seeds and only questions agreed by most QA models are selected. When minimal target-domain annotation is available, we can also learn to reweight pseudo samples based on the validation loss~\cite{sun2021few}, or use RL to select samples that lead to validation performance gains (value estimation)~\cite{yue2022synthetic}.

\paragraph{Discussion} Compared with SCT, QG generates natural sentences as pseudo questions, which makes it closer to the real use case. If the generator is properly trained, DR trained with QG can even match the fully-supervised performance while using zero in-domain annotations~\cite{wang2022gpl,ren2022thorough}. The biggest challenge lies in the training of the question generator. If the question distribution from our target domain is very different from available domains, the question generator might not be able to adapt well. A solution is to apply semi-supervised learning when target-domain questions are available. We can create pseudo training instances by back-training to adapt the question generator to the target domain~\cite{kulshreshtha2021back,shen2022product}. Another issue is the one-to-many mapping relations between questions and documents. Under this situation, seq2seq learning tends to produce safe questions with less diversity and high lexical overlap with the document. For example, \citet{shinoda2021can} found that QG reinforces the model bias towards high lexical overlap. We will need more sophisticated training techniques such as latent-variable models~\cite{yao2018teaching,lee2020generating,li2022generating} and reinforcement learning~\cite{yuan2017machine,zhang2019addressing} to alleviate the model bias towards safe questions.
\section{Leveraging Documents + Question}
\label{sec:questions}
This section assumes additional access to unlabeled questions $Q_t$ in the target domain. In practice, annotating question-document relations requires a time-consuming process of reading long documents, with domain knowledge and careful sampling strategies to ensure enough positive samples. Unlabeled questions are much easier to obtain either through real user-generated content or simulated annotations, so it is common to have a predominance of unlabeled questions in low-resource scenario~\footnote{If no real questions are available, we can use synthetic questions from question generation, then apply same techniques in this section~\cite{wang2022gpl,thakur2022domain}.}. To leverage a documents + questions resource, we discuss two popular techniques in this section: (1) distant supervision and (2) domain invariant learning. 
\subsection{Distant Supervision}
\label{sec:ds}

\emph{Distant supervision} applies different methods to automatically establish the missing relevance labels between questions and documents. These labels are then used as ``distant'' signals to supervise DR. Specifically, given a question-document pair $(q, d)$, we have a teacher $T$ to provide a pseudo label $T(q, d)$ for them. This pseudo label is used to train the DR. The training objective can be formulated as:
\begin{align}
\begin{split}
     \min_{R}\mathbb{E}_{q \in Q_t, d \in D_t} \mathcal{L}(R(q,d),T(q,d))
\end{split}
\end{align}
where $\mathcal{L}$ is the loss function that encourages similarity between $R(q,d)$ and $T(q,d)$. The success of distant supervision lies in choosing the proper teacher and the loss function, which we will cover in the following section.

\begin{table}
\centering
\begin{adjustbox}{max width=0.47\textwidth}
\small
\begin{tabular}{l|l}
\toprule
\multirow{4}{*}{Teacher} &  Sparse Retriever$\dagger$ \\
& Pretrained Language Model$\dagger$\\
& Large-sized Model$\ddagger$\\
& Multi-representation-encoder$\ddagger$\\
& Cross-encoder$\ddagger$\\
& Self$\ddagger$\\
\midrule
\multirow{5}{*}{Loss} &  Cross Entropy loss\\
& Contrastive loss\\
& Knowledge Distillation loss\\
& MarginMSE loss\\
& Hinge loss\\
\toprule
\end{tabular}
\end{adjustbox}
\caption{\label{tab:ds}
\small Different choices for distant supervision. $\dagger$ means unsupervised model. $\ddagger$ means supervised model (out-of-domain annotations can be used to train it).
}
\end{table}

\paragraph{Sparse Retriever (SR) as Teacher}
Recent research finds that DR and SR are complementary. DR is better at semantic matching while SR is better at capturing exact match and handling long documents~\cite{chen2021salient,luan2021sparse}. SR is also more robust across domains~\cite{thakur2021beir,chen2022out}. This motivates the use of unsupervised sparse retrievers (SRs) like BM25 as teachers. For example, \citet{dehghani2017neural,nie2018multi} train a neural model on training
examples that are annotated by BM25 as the weak supervision signal. \citet{xu2019passage} apply 4 scoring functions to auto-label the questions and documents: (1) BM25 score, (2) TF-IDF score, (3) cosine similarity
of universal embedding representation~\cite{cer2018universal} and (4) cosine similarity of the last hidden
layer activation of pretrained BERT model~\cite{kenton2019bert}.
Both papers observe that the resulting model outperforms BM25 itself
on the test data. \citet{chen2021salient} show distilling knowledge from BM25 helps the DR on matching rare entities and improves zeroshot out-of-domain performance.

\paragraph{PLM as teacher}
As PLMs already encode significant linguistic knowledge, there have also been attempts at using prompt-based PLMs as teachers to auto-label question-document pairs~\cite{smith2022language,zeng2022weakly}. We can use simple prompts like ``\textit{Please write a question based on this passage}'', concatenate the document and question, then use the probability assigned by the PLM to auto-label question-document pairs. To maximize the chances of finding positive documents, normally we first obtain a set of candidate documents by BM25, then apply PLM to auto-label the candidate set~\cite{sachan2022improving}. This can further exploit the latent knowledge inside PLMs that has been honed through pretraining, so it often shows better performance compared with distant supervision only using BM25~\cite{nogueira2020document,singh2022questions}.

\paragraph{Supervised Model as Teacher} Compared with the above two unsupervised teachers, a more common choice is to use a supervised model fine-tuned on labeled data as the teacher. When in-domain annotations  are  not  enough,  we  can  leverage  out-of-domain (OOD) annotations, if available, to train the teacher. The teacher often employs a more powerful architecture. It may not be directly applicable in downstream tasks due to the latency constraints, but can be useful in providing distant signals for training the DR. For example, previous research has shown that models with larger sizes or multi-representation/cross-encoder structures generalize much better on OOD data~\cite{pradeep2020h2oloo,lu2021less,ni2021large,rosa2022no,muennighoff2022sgpt,zhan2022evaluating}. After training a more powerful teacher on OOD annotations, applying the teacher to provide distant supervision through target-domain question and document collections can significantly improve the retrieval performance in low-resource scenarios~\cite{hofstatter2021efficiently,lin2021batch,lu2022ernie}. \citet{kim2022collective} further show that we can even use the same architecture for the teacher and student. They expand the query with centroids of word embeddings from top retrieved passages (using BM25), and then use the expanded query for self knowledge distillation. Similar ideas of applying the model itself as the teacher have also been explored by \citet{yu2021few,kulshreshtha2021back,zhuang2022characterbert}.

\paragraph{Choice of Loss Function}
The typical choice for the loss function is the contrastive loss or cross entropy loss as in training the standard DR. This treats the output from the teacher as one-hot targets and ignores the soft distribution of labels. To leverage richer knowledge from the continuous output distribution of the teacher, we can also use the knowledge distillation (KD) loss to minimize the KL divergence between the output of the teacher and the student DR~\cite{hinton2015distilling}. Different techniques have been proposed to improve the KD process. \citet{qu2021rocketqa} uses a confidence-based filter to improve the quality of training signals. \citet{zhou2022bert} applies meta learning to reduce the teacher-student gap. \citet{zeng2022curriculum} shows that curriculum learning can be used to speed up the learning of the student dense retrieval. Also commonly used are margin-based losses such as hinge loss~\cite{dehghani2017neural,xu2019passage} and MarginMSE loss~\cite{hofstatter2020improving,wang2022gpl}. Compared with cross entropy, contrastive and KD losses, margin-based losses are less likely to overfit the imperfect scores of the pseudo training data. If the output from the teacher model has significant noise, using margin-based losses are usually preferred and can lead to a more robust model.

\paragraph{Discussion} Distant supervision works directly on real questions and documents so that the model can be adapted to the target-domain data distribution. The bottleneck is the quality of the teacher model. If the output from the teacher model contains substantial noise, the resulting DR will also overfit to the noise and have poor performance. A popular noise-resistant algorithm used in distant supervision is confidence-based filtering~\cite{mukherjee2020uncertainty,yu2021fine}. However, this works only if the output score of the teacher is a reasonable estimate of the actual confidence. If the task is challenging, the output score becomes misleading and cannot be used to reduce the noise~\cite{zhu2022meta}. Another problem is the computational efficiency. When applying more powerful models as the teacher, the model inference can be significantly slower~\cite{ni2021large}. Obtaining the supervision labels for all training samples can be very time-consuming~\cite{xu2020computation,rao2022parameter}.

\subsection{Domain-invariant Learning}
\label{sec:dat}
 \emph{Domain-invariant learning} (DiL) aims to learn domain invariant representations to facilitate the generalization from source to target domain~\cite{ganin2015unsupervised,ganin2016domain}. DiL approaches DR training with a two-component training objective: (1) a standard ranking loss applied to OOD question-document annotations, and (2) a DiL loss to minimize the discrepancy between the representations of the source and target domain. The ranking loss $\mathcal{L}_{rank}$ is defined over the OOD data as:
\begin{align}
\begin{split}
     -\min_{R}\mathbb{E}_{q,d^{+}, d_{1\sim n}^{-}\in Q_o \times D_o}\mathcal{O}(R,q,d^{+},d_{1\sim n}^{-})
\end{split}
\end{align}
The DiL objective can be formulated as:
\begin{align}
\begin{split}
     &\mathcal{L}_{DiL}=\min_{E1,E2}\mathbb{E}_{q_t \in Q_{t}, d_t \in D_{t}, q_o \in Q_o, d_o \in D_o}\\
     &\mathcal{D}(E1(q_o)||E1(q_t)) + \mathcal{D}(E2(d_o)||E2(d_t))
\end{split}
\end{align}
where $\mathcal{D}$ is a function to measure the discrepancy. $E_1$ and $E_2$ are the question and document encoder respectively. The final loss is formed by combining both loss objectives:
\begin{align}
\begin{split}
     \mathcal{L}=\mathcal{L}_{rank} + \lambda \mathcal{L}_{DiL}
\end{split}
\end{align}
where $\lambda$ is an justable hyperparameter.
There are two popular ways of measuring the discrepancy: (1) maximum-mean discrepancy and (2) adversarial learning.
\paragraph{Maximum-Mean Discrepancy (MMD)}
\emph{MMD} is a distance on the space of probability
measures by computing the norm of the difference between two domain means~\cite{quinonero2008covariate} and has been a popular technique to learn domain-invariant representations~\cite{tzeng2014deep,long2015learning}. Its definition is:
\begin{align}
\begin{split}
     \mathcal{D}_{MMD}(P,Q)=||\mu_P-\mu_Q||_{\mathcal{H}}
\end{split}
\end{align}
where $P$ and $Q$ are two distributions, $\mu$ is the mean of the distribution and $\mathcal{H}$ refers to the corresponding Reproducing Kernel
Hilbert Space~\cite{berlinet2011reproducing}. The mean is usually estimated by the empirical mean over a batch.

\citet{tran2019domain} applies MMD to domain adaptation for information retrieval and estimate the MMD with first-order mean vectors. ~\citet{chen2020esam,liu2022collaborative} further add a regularization to minimize second-order statistics
(covariances) of the source and target features. Although sharing the
same mean and covariance matrix is not sufficient to guarantee domain-invariant distributions, \citet{tran2019domain} experimentally find that projecting the distributions
of representations in random directions produced one-dimensional
distributions that appeared Gaussian, so that minimizing the first two moments suffices to guarantee the equivalence of two distributions.

\paragraph{Adversarial Learning}
\emph{Adversarial Learning} estimates density ratios of two distributions through a discriminator $f$~\cite{goodfellow2014generative}. By modifying the training objective of $f$, it can be used to estimate different distance metrics between two distributions such as Jesson-Shannon divergence and wasserstein distance~\cite{goodfellow2016nips,arjovsky2017wasserstein}. In the standard setting,
$f$ is trained by:
\begin{align}
\begin{split}
     \min_{f}\mathbb{E}_{x \sim P/Q}-\log f(l_x|x)
\end{split}
\label{eq: domain-classifier}
\end{align}
where $l_x$ is the label to indicate if $x$ comes from the distribution $P$ or $Q$. The discrepancy is defined as:
\begin{align}
\begin{split}
     \mathcal{D}_{adv}(P,Q)=\mathbb{E}_{x \sim P/Q}\mathcal{L}(\mathcal{U}|f(l_x|x))
\end{split}
\label{eq: adversarial}
\end{align}
where $\mathcal{U}$ is the uniform distribution and $\mathcal{L}$ is the loss function. Common choices for $\mathcal{L}$ is the cross entropy or KL divergence loss. The model is trained to encode the questions and documents from different domains into the same distribution space so that $f$ cannot distinguish from which domain they came. Eq~\ref{eq: domain-classifier} and Eq~\ref{eq: adversarial} are trained iteratively to learn from each other.

\citet{lee2019domain} apply adversarial learning on question-answering and show advantages over baseline models. \citet{wang2019adversarial} show adversarial learning can also be applied to synthetic questions generated from the target-domain passages, avoiding the need to obtaining large amounts of target-domain questions. \citet{shrivastava2022qagan} proposed an annealing trick to prevent the discriminator from providing misleading signals in early training stages. All the above focus on training the reader component. \cite{cohen2018cross} apply the adversarial loss to regularize the standard ranking tasks. \citet{xin2022zero} is the only work that applies DiL on DR training. To balance the efficiency and robustness of
the domain classifier learning, they propose maintaining a momentum queue to average over past batches, so that we do not need to feed all data points to the classifier at every batch. 

\paragraph{Discussion} Though impressive improvement is observed, DiL requires careful parameter tuning to balance the ranking loss and the DiL loss. When applying adversarial learning as the discrepancy measure, it can be especially challenging to adjust the update of the discriminator and the DR~\cite{wang2020cross,shrivastava2022qagan}. The training process can be much more difficult compared to other approaches. Furthermore, DiL requires both target-domain questions and documents. \citet{xin2022zero} verified that it does improve over the simplest baseline regarding the domain adaptation capability, but they did not compare it with other techniques like self-contrastive learning or question generation that only require target-domain document collections. It is not clear to which extent DAT is helpful when combined with other techniques or how practical is to apply it considering the cost and difficulty of training with the DiL objective.

\section{Leveraging Documents + QA Pairs}
\label{sec:qa_pairs}
Many domains have large numbers of already answered questions from customer services, technical support or web forums~\cite{huber2021ccqa}. These question-answer pairs can provide richer information than only unlabeled questions~\footnote{If no real QA pairs are available, we can use heuristics like masked salient entities~\cite{guu2020retrieval} to form pseudo pairs, then apply the same training techniques in this section.}. However, most answers are based on personal knowledge, derived from experience, and do not include a reference to any external documents. This prevents its direct use as training data for the retriever. This section introduces three mainstream techniques to exploit QA pairs despite this difficulty: (1) direct fine-tuning, (2) answer-document mapping and (3) latent-variable model.

\subsection{Direct Fine-tuning}
\label{sec:df}
 \emph{Direct Fine-tuning} is the most straightforward way to leverage QA pairs. It directly fine-tunes the document retriever on QA pairs and does not distinguish between documents and answers~\cite{lai2018supervised}. The training objective is:
\begin{align}
\begin{split}
     \max_{R}\mathbb{E}_{q,a^{+}, a_{1\sim n}^{-}\in Q_t\times A_t}
     \mathcal{O}(R,q,a^{+},a_{1\sim n}^{-})
\end{split}
\end{align}
where $(q,a^+)\in Q_t \times A_t$ are question-answer pairs in the target domain, $a_{1\sim n}^{-}$ are sampled $n$ negative answers and $\mathcal{O}$ is the standard ranking objective.

\paragraph{Discussion}
Direct fine-tuning on QA pairs has been a common practice to ``warm up'' the DR. For large-sized models, this can be crucial to fully leverage the model capacity since we often have orders of magnitude more QA pairs than question-document annotations~\cite{ni2021large,ouguz2021domain}.
However, the style, structure and format differ between the documents and the answer~\cite{shen2022product}. The answer is a direct response to the question, and so it is easier to predict due to its strong semantic correlation with the question. Whereas the document can be implicit and may contain fewer obvious clues that can imply an answer; deep text understanding is required to predict the relevance between questions and documents~\cite{zhao2021distantly,shen2022semipqa}. Therefore, direct fine-tuning can be a good strategy to get a decent starting point for the DR, but it may be insufficient to reach satisfying results as a standalone technique.

\subsection{Answer-Document Mapping}
\label{sec:adm}
 Answer-Document Mapping leverages a mapping function to automatically link answers to the corresponding documents. The training objective is:
\begin{align*}
\begin{split}
     \max_{R}\mathbb{E}_{q, a \in (Q_t,A_t), d_{1\sim n}^{-} \sim D_t}\mathcal{O}(R,q,M(a),d_{1\sim n}^{-})
\end{split}
\end{align*}
where $(q,a)\in (Q_t,A_t)$ are question-answer pairs in the target domain, $M$ is a mapping function from an answer to its corresponding document, and $\mathcal{O}$ is the standard ranking objective.

\paragraph{Choices of Mapping Function} The mapping function is based on hand-crafted heuristics. 
For long-form descriptive answers, a popular way is to map them to documents with the highest ROUGE score~\cite{lin2004rouge} since the answers can be considered as summaries of the original documents~\cite{fan2019eli5}. For short-span answers, a popular way is to map them to the top-ranked documents retrieved from BM25 that contain the answer span~\cite{karpukhin2020dense,sachan2021end,christmann2022conversational}.

\paragraph{Discussion} Answer-document mapping has been widely adopted to construct large-scale datasets for information retrieval~\cite{joshi2017triviaqa,dunn2017searchqa,elgohary2018dataset}. Compared with direct fine-tuning, answer-document mapping trains model on question-document pairs not question-answer pairs so that it is closer to our real objective. However, there are potential errors if automatic mapping is not accurate. For example, frequent answers or entities might lead to false positive mappings. It is also difficult to find positive documents with little lexical overlap with the answer~\cite{izacard2021distilling}. Models trained only with Answer-Document-Mapping can easily overfit to the bias of the mapping function and fail to find the true relevance between questions and documents~\cite{du2022pregan}.

\subsection{Latent-variable Model}
\label{sec:lvm}
 Similar to answer-document mapping, approaches based on \emph{Latent-variable model} also train the retriever on question-document pairs. However, instead of relying on a heuristic-based mapping function, the missing link between documents and answers is considered as a ``latent variable'' within a probabilistic generative process~\cite{brown1993mathematics}. The retriever $R$ and reader $G$ are jointly trained to maximize the marginal likelihood:
\begin{align}
\begin{split}
\max_{R,G}\mathbb{E}_{q,a \in (Q_t, A_t)}\log \sum_{z\sim Z}{R}(z|q){G}(a|q,z)  
\end{split}
\label{eq:latent}
\end{align}
where $Z$ indicates all possible document combinations. Directly optimizing over Eq~\ref{eq:latent} is infeasible as it requires enumerating over all documents. A closed-form solution does not exist due to the deep neural network parameterization of $R$ and $G$. The following section explains popular options in performing the optimization. An overview can be seen in Table~\ref{tab:lvm}.

\begin{table}
\centering
\begin{adjustbox}{max width=0.47\textwidth}
\small
\begin{tabular}{l|l}
\toprule
\textbf{Distribution of $R(z|q)$} & \textbf{Optimization Method}\\
\midrule
Categorical & Top-k Approximation\\
\midrule
\multirow{2}{*}{Multinomial} &  EM-Algorithm \\
& Learning from Reader\\
\toprule
\multicolumn{2}{c}{\textbf{Objective}:$\max_{R,G}\mathbb{E}_{q,a \sim (Q_t, A_t)}$}\\
\multicolumn{2}{c}{$\log \sum_{z\sim Z}{R}(z|q){G}(a|q,z)$}\\
\end{tabular}
\end{adjustbox}
\caption{\label{tab:lvm}
\small Different ways of optimizing the latent-variable models that leverage documents and question-answer pairs.
}
\end{table}

\paragraph{Top-K Approximation}
A popular way is to assume a categorical distribution for ${R}(Z|q_t)$, that is to assume for each question only a single document is selected and the answer is generated from that one document. Eq~\ref{eq:latent} can be approximated by enumerating over only the top-k documents, assuming the remaining documents having negligibly small contributions to the likelihood:
\begin{align*}
\begin{split}
\max_{R,G}\mathbb{E}_{q,a \in (Q_t, A_t)}\log \sum_{z\sim D_{topk}}{R}(z|q){G}(a|q,z)  
\end{split}
\label{eq:individual-topk}
\end{align*}
This has been a popular choice in end-to-end retriever training~\cite{lee2019latent,guu2020retrieval,lewis2020retrieval,shuster2021retrieval,ferguson2022retrieval}. Despite its simplicity, the top-k approximation training has two main drawbacks. (1) The approximation is performed on the top-k documents obtained from the retriever. If the retriever is very weak at the beginning of training, the top-k documents can be a bad approximation and the model might struggle to converge~\cite{shen2019select,shen2019improving}. (2) The assumption that documents follow a categorical distribution might be problematic especially if the answer requires evidence from multiple documents~\cite{zhong2019coarse,wang2022deep}.

\paragraph{EM Algorithm}
\emph{EM Algorithm}~\cite{dempster1977maximum} can be applied to address the second drawback of \emph{Top-K Approximation}. Instead of using the categorical distribution, we can assume a multinomial distribution for ${R}(Z|q)$. Multiple documents can be selected at the same time and an answer can be generated from multiple documents. The cost of this relaxation is the increased difficulty of optimization. Approximating the joint likelihood from top-$k$ samples becomes infeasible due to the combinatorial distribution of documents. \citet{singh2021end} propose optimization it with the EM algorithm under an independent assumption about the posterior distribution of ${R}(z|q)$:
\begin{equation}
\label{eq:joint_k}
\begin{split}
&\max_{R,G}\mathbb{E}_{q,a \in (Q_t, A_t)}[\log\sum_{z\in D_{topk}}{R}(z|q)\\
&\times SG({G}(a|q,z))+\log {G}(a|q,D_{topk})]
\end{split}
\end{equation}
where $D_{topk}$ are the top-$k$ documents obtained with the retriever and $SG$ means stop-gradient (gradient is not backpropagated through). Similarly, \citet{zhao2021distantly} applies the hard-EM algorithm (Viterbi training)~\cite{allahverdyan2011comparative} to train the retriever, which only treats the document with the highest likelihood estimated by the reader as positive. \citet{singh2021end} also find that Eq~\ref{eq:joint_k} is quite robust with respect to parameter initialization, and that it can be optimized without sophisticated unsupervised pretraining methods.

In Eq~\ref{eq:joint_k}, the training signal for the retriever essentially degenerates to the same as in the \emph{Top-K Approximation} case, except that the reader conditions on all top-k documents to generate the answer. It is unclear how much any improvement can be attributed to the reader or the retriever. \citet{sachan2021end} find that conditioning on top-k documents has better end-to-end performance but worse retrieval performance (though they use a different training signal than Eq~\ref{eq:joint_k} for the retriever), suggesting its benefit could be only a more powerful reader. Ffiner-grained methods may be needed to train the retriever properly.

\paragraph{Learning from Reader}
Another way to optimize the retriever in Eq~\ref{eq:latent} is to leverage attention scores from $G$. The assumption is that when training the reader to generate the answer, its attention score is a good approximation of question-document relevance. The training objective is:
\begin{align}
\begin{split}
\min_{R,G}&\mathbb{E}_{q,a \in (Q_t, A_t)} \sum_{z\sim D_{topk}}\mathcal{L}(A_z|R(z|q))\\
&-\log {G}(a|q,Z=D_{topk})  
\end{split}
\end{align}
where $G$ is trained to generate the right answer based on the question and the top-k documents, same as in the EM algorithm. $A_z$ is the attention score of $G$ on the document $z$. $\mathcal{L}$ is the loss function to encourage the similarity between distributions of the attention scores and retrieving scores.

\citet{izacard2021distilling} propose a training process that optimizes $R$ and $G$ iteratively. $R$ is trained to minimize KL divergence between relevance and attention scores. \cite{lee2021you} propose to jointly optimize $R$ and $G$ and apply a stop-gradient operation on $G$ when training the $R$. \citet{sachan2021end} use retriever scores to bias attention scores. These can be considered as first-order Taylor
series approximations of Eq.~\ref{eq:latent} by replacing ${R}(Z|q_t)$ with attention scores~\cite{deng2018latent}. However, the gap between this approximation and the original objective can be large when the attention score has high entropy~\cite{ma2017dropout}. As noticed in \citet{yang2020retriever}, the relevance score from $R$ and the attention score from $G$ can be very different. Purely training the retriever from the reader's attention scores could be suboptimal.

\paragraph{Discussion} If properly trained, latent-variable models can perform close to fully supervised models~\cite{zhao2021distantly,sachan2021end}. The main challenge is the training difficulty. In practice, we can often initialize the retriever with \emph{direct fine-tuning} or \emph{answer-document mapping} to make the training more stable. If the independence assumption made by Eq~\ref{eq:joint_k} does not hold, we need to resort to more complex optimization algorithms. A potential direction is to apply a Dirichlet prior over ${R}(z|q_t)$, which is a conjugate distribution to the multinomial distribution~\cite{minka2000estimating}, with the result that the sampled documents are not independent individuals but a combination set. Eq~\ref{eq:latent}
can then be estimated by rejection sampling~\cite{deng2018latent} or a Laplace approximation~\cite{srivastava2017autoencoding} so as to avoid the independence assumption about the posterior distribution. Nonetheless, this will further increase the training complexity, which is already a key bottleneck for training the DR.

\section{Other Approaches}
\label{sec:others}
There are also promising research directions that do not require any of the above-mentioned in-domain resources. We provide a brief overview oin this section.
\paragraph{Integration with Sparse Retrieval (SR)}
As mentioned in~\cref{sec:ds}, recent research finds that DR and SR are complementary~\cite{chen2021salient,luan2021sparse}.
Integrating scores can potentially combine the strengths of both~\cite{wang2021bert}. In the low-resource scenario, the integration can be especially beneficial. For example, \citet{ram2021learning,ren2022thorough} show interpolating the DR score with BM25 can further improve the performance of unsupervised retrievers. \citet{kuzi2020leveraging} use RM3~\cite{abdul2004umass} built on the top lexical results to
select deep retriever results as the final list. \citet{chen2022out} integrate DR and SR with a more robust rank-based score and shows significant improvement on zeroshot evaluations.

\paragraph{Multi-Task learning}
Multi-task learning is a popular method to improve model generalization~\cite{aghajanyan2021muppet}. By fine-tuning on a mixture of various tasks, the model is able to learn different kinds of knowledge with less tendency to overfit to one specific task. OOD annotations can be optionally mixed into the tasks. ~\citet{maillard2021multi} fine-tunes a universal retriever on 8 different retrieval tasks and finds it is beneficial not only in the fewshot scenatio, but also when abundant in-domain annotations are available. \citet{li2021-multi-task-dense} proposes to train individual dense retrievers for different tasks and aggregate their predictions during test time. This can reduce the problem of corpus inconsistency when fine-tuning a single model on a mixture of all corpora. This can be further combined with meta learning, which adopts a ``learning to learn'' paradigm to quickly adapt the model to different domains~\cite{finn2017model,qian2019domain,dai2020learning,park2021unsupervised}.

\paragraph{Parameter Efficient Learning}
Parameter efficient Learning methods like
adapters~\cite{houlsby2019parameter} and prompt-tuning~\cite{li2021prefix,liu2021gpt} have been widely adopted in many NLP tasks. By modifying only a subset of the model parameters, the model is able to retain most general knowledge it learnt in the pretraining stage. Such models often generalizes better when only limited annotations are available. The approach can be used to fine-tune the model on OOD as well as in-domain annotations, if available. ~\citet{tam2022parameter} applies parameter-efficient prompt tuning for text retrieval across in-domain, cross-domain, and
cross-topic settings. By fine-tuning only $0.1\%$ of model parameters, they find better generalization
performance than methods in which all parameters are updated.

\section{Conclusions and Open Issues}
\label{sec:conclusions}
In this survey we review mainstream techniques for training dense retrievers in open-domain question answering under low-resource scenarios and provide a structured way of classifying them according to the required resource. For each technique we discuss different options in their sub-components and summarize the pros and cons. The survey can serve as a reference book for practitioners to identify the techniques suitable under specific low-resource scenarios. There are still quite a few open issues left. We have pointed to main challenges for each technique in corresponding sections. As a final wrap-up, we list general problems that we believe worth exploring for future research:
\begin{itemize}

\item \emph{How to select the most suitable techniques?} Despite the wide range of applicable techniques, it is non-trivial to decide how to select the best one in approaching a new domain. In practice, the decision is usually made based on empirical experiments for each individual domain, but a systematic approach is largely missing.

\item \emph{To which extent are these techniques complementary?} Existing works normally compares performance only to similar types of methods but not across the range of techniques and resources. This makes it hard to see where the different approaches complement each other and how they can be combined effectively.

\item \emph{Do methods work across languages?} The vast majority of current research is conducted on English datasets. Even though all described methods in this survey have no hard restriction on their language, it is possible that performance would vary across languages, especially for methods that rely on handcrafted heuristics.

\end{itemize}
Answering these questions is not easy given the different requirements of data. For most non-English languages, it is hard to obtain large amounts of resources across domains. Future research must address the development of both modelling and data collection.

\bibliography{custom}
\bibliographystyle{acl_natbib}

\end{document}